\documentclass{ifacconf}

\usepackage{graphicx}      % include this line if your document contains figures
\usepackage[numbers]{natbib}        % required for bibliography
\usepackage{amsmath} 
\usepackage{amssymb}  
\usepackage{color}
\usepackage{enumerate}

\makeatletter
\let\old@ssect\@ssect % Store how ifacconf defines \@ssect
\makeatother
\usepackage{hyperref}

\makeatletter
\def\@ssect#1#2#3#4#5#6{%
  \NR@gettitle{#6}% Insert key \nameref title grab
  \old@ssect{#1}{#2}{#3}{#4}{#5}{#6}% Restore ifacconf's \@ssect
}
\makeatother
\usepackage{multirow}
\usepackage{mathtools}
\usepackage{booktabs}
\usepackage{epstopdf}
\usepackage{algcompatible}
\usepackage[normalem]{ulem}
\usepackage{subcaption}
\usepackage{url}
\usepackage{bm}
\usepackage{caption}
\usepackage{balance}
\theoremstyle{definition}

\theoremstyle{remark}

\newcommand{\norm}[1]{\left\lVert#1\right\rVert}
\theoremstyle{assumption}
\newtheorem{assumption}{Assumption}
\begin{document}
\begin{frontmatter}

\title{Data-Driven Optimization for Deposition with Degradable Tools} 

\thanks[footnoteinfo]{This work was supported by ONR-N00014-18-1-2833, NSF-1931853, and AFRI Competitive Grant no. 2020-67021-32855/project accession no. 1024262 from the USDA National Institute of Food and Agriculture. This grant is being administered through AIFS: the AI Institute for Next Generation Food Systems (\href{https://aifs.ucdavis.edu}{https://aifs.ucdavis.edu}).}

\author[First]{Tony Zheng},
\author[First]{Monimoy Bujarbaruah}, 
\author[First]{Francesco Borrelli} 

\address[First]{University of California, Berkeley, CA 94720 USA (e-mail:\{tony\_zheng, monimoyb, fborrelli\}@berkeley.edu).}  

\begin{abstract}                % Abstract of not more than 250 words.
We present a data-driven optimization approach for robotic controlled deposition with a degradable tool. Existing methods make the assumption that the tool tip is not changing or is replaced frequently. Errors can accumulate over time as the tool wears away and this leads to poor performance in the case where the tool degradation is unaccounted for during deposition. In the proposed approach, we utilize visual and force feedback to update the unknown model parameters of our tool-tip. Subsequently, we solve a constrained finite time optimal control problem for tracking a reference deposition profile, where our robot plans with the learned tool degradation dynamics. 
We focus on a robotic drawing problem as an illustrative example.  Using real-world experiments, we show that the error in target vs actual deposition decreases when learned degradation models are used in the control design.
\end{abstract}

\begin{keyword}
Autonomous robotics systems, Data-driven optimal control, Modeling.
\end{keyword}

\end{frontmatter}
%===============================================================================

\section{Introduction}
Robotic manipulation in contact-rich tasks have seen great advancements in recent years \citep{Suomalainen2022}. There has been a push towards robots that can help in daily household chores such as folding clothes \citep{Goldberg2022}, wiping surfaces \citep{Leidner2016}, or various kitchen tasks \citep{Ebert2022}. While these tasks are certainly challenging, there are still unaddressed problems in the field where the contact tool itself changes over time. Examples include cutting blades that decrease in sharpness through repeated use, sandpaper which wears away, or chalk for marking surfaces. 

For this work, we consider the deposition with degradable tools in the application of robotic drawing. Artwork and videos generated by AI in recent works \citep{Ramesh2021, Dalle2, phenaki,Rombach2022,Saharia2022} have been able to produce complex creations that could easily be mistaken as drawn by professional artists. Text-to-image generation has been able to produce some incredible results that allow the user to create highly specific combinations of subjects performing actions in locations, even in the artwork style that they desire \citep{Dalle2,Rombach2022,Saharia2022}. Research in predictive language models like GPT-3 \citep{Brown2020} linked with large image databases \citep{imagenet} have been a huge part of these advancements. However, one major hurdle has been translating these artworks into the real world. There are many complex physical interactions involved in various mediums of artwork such as oil painting, pencil sketching, water-colors, etc. Work has been done the decomposition of images into individual strokes to reproduce the image \citep{Huang2019,Tong2021}. The final drawn images can be highly accurate but they lack insight in actually making a robot hold a paintbrush or pencil to produce those strokes. Current state-of-the-art approaches for robotic drawing are predominately hand-tuned open loop sequences with custom end-effectors that make it easier to have a constant force output and frequent reset sequences to allow for consistency\citep{Jain2015,Kotani2019}. A more accurate replication of human drawing would take the deformation of a tool-tip into account and change the policy accordingly. To the best of our knowledge, there has not been works that use visual feedback to update the model of a tip's degradation to produce more accurate strokes.
 \begin{figure}[h!]
	\centering 	\includegraphics[width=0.75\columnwidth]{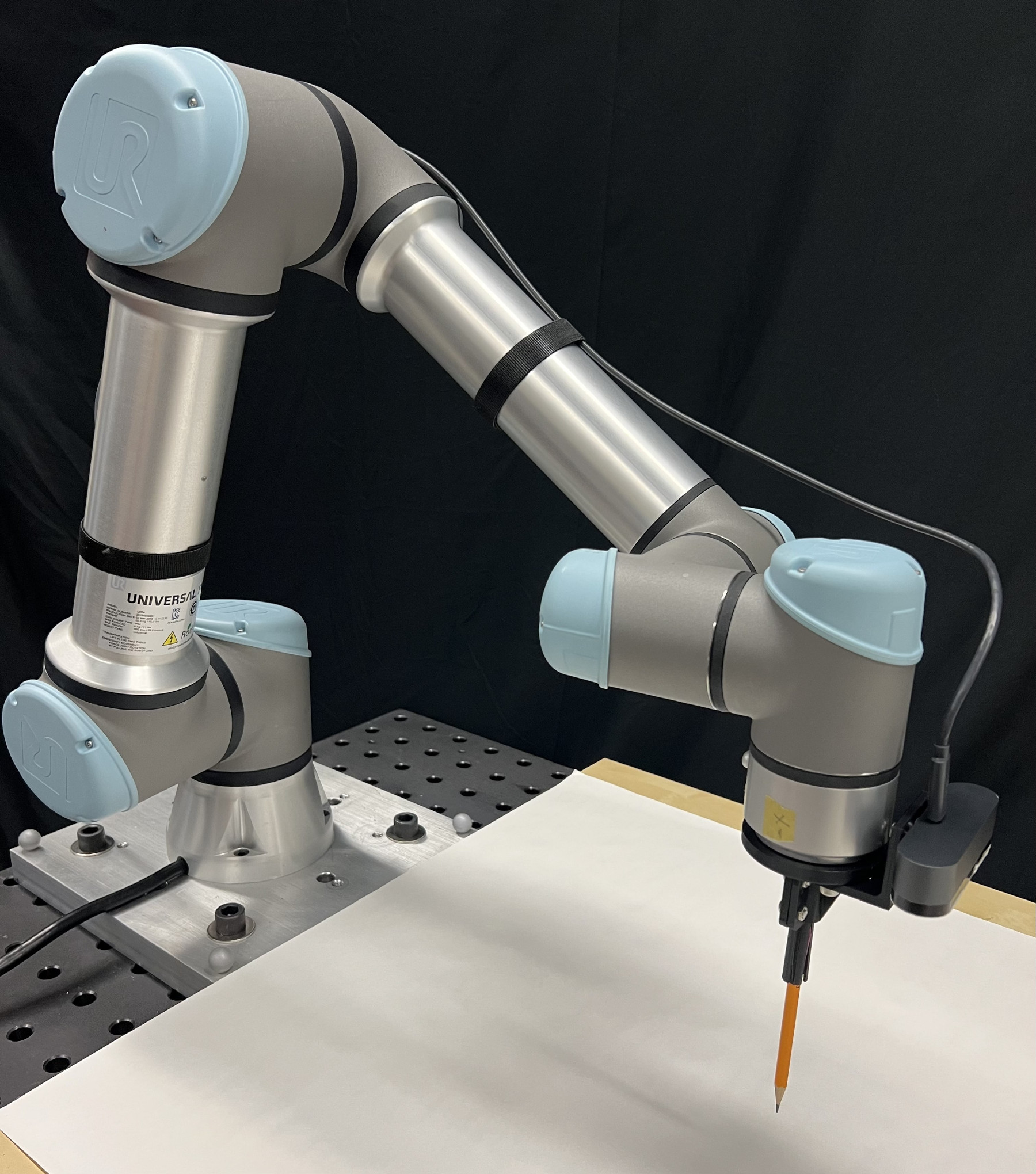}
	\caption{The considered experimental setup.}
	\label{fig:exp_space}    
\end{figure}

In this paper, we formulate the deposition task as a model-based constrained finite time optimal control problem, where we model the deposition and the degradation of the tool tip. The parameters of these models are not known a-priori, and we learn these using collected data. We focus on the specific example of a robot sketching using a pencil. The unknown degradation and deposition models of the pencil are parameterized as a function of the applied force and the distance drawn. We present detailed experiments with a UR5-e robot where we show that accounting for the degradation of the tip and planning strokes accordingly improves the sketching quality measured in terms of the difference in stroke width error.

\section{Related Work}
 \subsubsection{Simulated stroke generation} Reinforcement learning approaches have been shown to work in generating realistic strokes. \citet{Xie2013} formulates brush strokes as an Markov decision process and solves it using policy gradient methods. \citet{Huang2019} takes target images and decomposes them into stroke sequences with deep reinforcement learning. \citet{Cao2019} introduced a CNN-based auto-encoder to generate multi-class sketches. \citet{Tong2021} uses the gradient of grayscale values to determine pencil stroke sequences used to recreate images. While these works produce strokes that appear similar to real ones, they are not tested on real robots where the actual deposition will likely differ from their expected. 
 \subsubsection{Real-world stroke generation}
 \citet{Wang2020} formulate calligraphy writing as a trajectory optimization problem and developed a novel dynamic brush model. They produce open-loop trajectories for that a robot can follow but do not close the loop. \citet{Lam2009} use linear regression to relate brush pressure with actual deposition width and generates trajectories to fill a desired calligraphy image. \citet{Bidgoli2020} trains a generative model on expert artist motion data in order to extract artistic style into robotic painting. They did not include closed-loop control and studied whether a playback of the artist motions with a robotic arm could produce brushstrokes similar to humans. \citet{Kotani2019} trains RNNs combined with LSTMs on images and produces commands for the robot that aim to draw strokes in continuous fluid motions. \citet{Gao2020} uses CNNs and GANs to extract sketch outlines from images and replicates the contours with a brush-pen. These actual robot demonstrations are done with open-loop sequences as well. 
 
 \citet{Adamik2022} explores pencil drawing and uses genetic algorithms to produce line segment sequences that can reproduce detailed drawings. Their robot uses a passive flexible tool to hold their custom graphite writing implement and compensates for drawing pressure changes over the surface. The tool-tip is designed in such a way that the graphite can be reset and calibrated frequently. \citet{Song2018} uses impedance control to draw on arbitrary surfaces using a pen. \citet{Dowd2021} corresponds force values with grayscale values of an image to shade using a pencil. These methods do not consider degradation of the tool-tip over time.

\section{Problem Formulation}
In this section, we describe the models used for our optimization based deposition problem. 

\subsection{Reference Stroke}
\begin{figure}[h!]
	\centering 	\includegraphics[width=0.7\columnwidth]{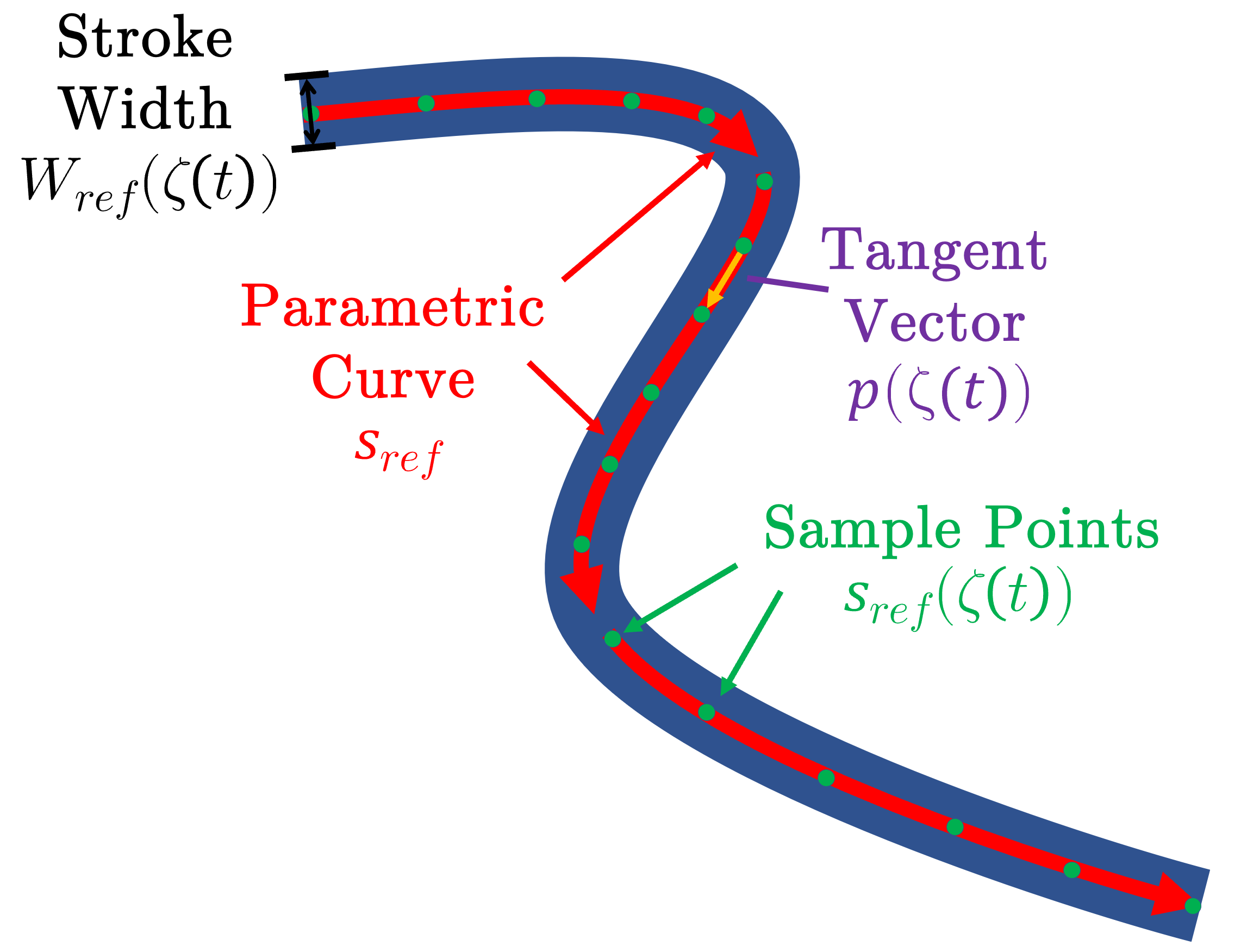}
	\caption{Stroke Analysis}
	\label{fig:stroke_width}    
\end{figure}

When drawing a picture, there are a limitless number of ways to decompose a desired image into individual strokes. We start with the assumption that each reference stroke is already given using stroke generation methods such as \citep{Tong2021, Huang2019}. The output of these generators are the parametric curve equations, width along the stroke, and color values. We convert these image coordinates so that the reference states, $s_\mathrm{ref}(\zeta(t))$, to be used in the cost function for our optimization problem are:
\begin{equation}\label{eq:parametric}
    s_\mathrm{ref}(\zeta(t)) = \begin{bmatrix} x_\mathrm{ref}(\zeta(t))\\y_\mathrm{ref}(\zeta(t))\\W_\mathrm{ref}(\zeta(t))
\end{bmatrix}, 
\end{equation}
where $t$ is the time step, $\zeta(t)$ is a parameter used to define the curve $s_\mathrm{ref}$, $\{x_\mathrm{ref}(\zeta(t)),y_\mathrm{ref}(\zeta(t))\}$ are the $x$ and $y$ positions (m) respectively and $W_\mathrm{ref}(\zeta(t))$ is the deposition width (m). Visually represented in Fig.~\ref{fig:stroke_width}, $p(\zeta(t))$ is the tangent vector to the stroke reference path $\{x_\mathrm{ref}(\zeta(t)),y_\mathrm{ref}(\zeta(t))\}$ sampled at any time step. Then, the deposition width $W_\mathrm{ref}(\zeta(t))$ is defined as the thickness of the stroke cross section at that point measured in a direction perpendicular to $p(\zeta(t))$. Henceforth, we will replace $(\zeta(t))$ with $(t)$ for the simplicity of notations. 

\subsection{End Effector Modeling}
We start the modelling of the discretized system beginning with the robot arm. We treat the end-effector of the robot arm as a single integrator:
%  \begin{equation}
%         \begin{aligned}
%     \bar{x}(t+1) = 
%     \bar{x}(t)
%      +
%      \Delta \bar{x}(t)
% \end{aligned} 
% \end{equation}
 \begin{equation}\label{eq:ee_model}
        \begin{aligned}
    \bar{x}(t+1) = 
    A\bar{x}(t)
     +
     Bu(t),
\end{aligned} 
\end{equation}
with $A = I_4$, $B = dt \cdot I_4$ where $I_n$ denotes the identity matrix of size $n$, and sampling time $dt = 0.008$s. The states and inputs are
\begin{equation}
\bar{x}(t) = 
\begin{bmatrix} x(t)\\y(t)\\z(t) \\ \psi(t)
\end{bmatrix}, 
u(t) = 
\begin{bmatrix} v_x(t)\\v_y(t)\\v_z(t)\\ \omega_{\psi}(t)
\end{bmatrix},
s(t) = 
\begin{bmatrix} x(t)\\y(t)\\W(t) 
\end{bmatrix},
\end{equation}
where $\{x(t), y(t), z(t)\}$ are end effector positions. The angle $\psi(t)$ is the angle between the pencil's ellipsoidal cross section's minor axis projected on the $xy$ plane, and the tangent to the stroke at time $t$. The inputs are their respective velocities. During control, we impose state and input constraints for the end-effector at all time steps $t \geq 0$ as given by:
\begin{align}\label{eq:constraints}
	x(t) \in \mathcal{X},~ u(t) 
	\in \mathcal{U},
\end{align}
for all $t = 0,1,\dots,N$, where $N > 0$ is the task horizon, and the sets $\{\mathcal{X},\mathcal{U}\}$ are polytopes. 

% \begin{assumption}\label{assump:tool_perp}
% \textcolor{black}{The end-effector is oriented such that the tool is perpendicular to the work surface and stays this way throughout the duration of the task.}
% \end{assumption}
As the tool is pushed deeper along the $z$-axis into the surface, a greater force is applied which we model in the following section.
\subsection{Force Map Modeling}
The force applied by the tool to the work surface can be modeled as a general nonlinear function 
\begin{equation}
F(t) = f_F(z(t) - z_\mathrm{ref}(t), \eta(t)),
\end{equation} 
where $z_\mathrm{ref}$ is the work surface height, $z(t) - z_\mathrm{ref}(t)\leq 0$ ensures that the tool is penetrating into the surface (for a downward direction), and $\eta(t)$ is a parameter that describes other contact conditions such as soft or dissipative contact. To obtain this relationship, we control the robot such that the tool tip makes contact with the work surface and then applies various penetration depths while taking force measurements. Because the actual function $f_F(\cdot, \cdot)$ is unknown for our tool, we fit a simplified linear model of the form shown below, ignoring the dependence on $\eta(t)$:
\begin{align}\label{eq:force_model}
F(t) = \theta(z(t) - z_\mathrm{ref}(t)) + \theta_0,
\end{align}
where parameters $\theta,\theta_0$ are unknown and to be learned from collected data. 

For drawing tools such as pencils, the contact force has a large effect on the rate at which the tip degrades. Our main contribution is that we utilize tip degradation of a rotated tip into the trajectory planning which we model in the next section.

\subsection{Tool Tip Modeling}
\begin{figure}[h!]
	\centering 	
	\includegraphics[width=\columnwidth]{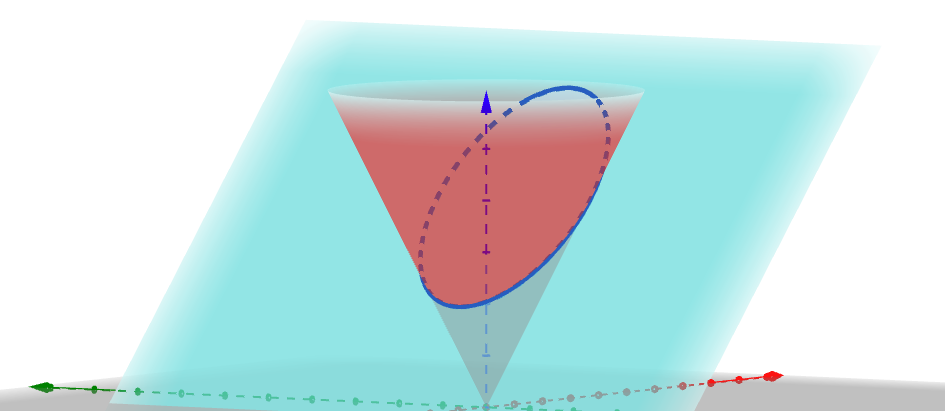}
	\caption{Intersection of a cone and hyperplane.}
	\label{fig:cone}    
\end{figure}

First, we model the pencil tip as a cone which will come into contact with a planar surface as shown in Fig.~\ref{fig:cone}.
This results in the general equation for the ellipse:
\begin{equation} \label{eq:ellipse}
 \frac{(m^2-a^2)^2(x+\frac{ad}{a^2-m^2})^2}{m^2d^2}+\frac{(m^2-a^2)y^2}{d^2}=1,
\end{equation} 
where $m$ is the slope of the cone, $a$ is the slope of the hyperplane, and $d$ is the vertical offset of the hyperplane from the cone's tip. 
\begin{assumption}\label{assump:tool_perp}
The slope ($a$) remains the same throughout the task duration to ensure that the shape of the surface stays an ellipse. This also relies on an upper bound for $d$ such that it does not make the hyperplane intersect past the top of the cone.
\end{assumption}
Under Assumption~\ref{assump:tool_perp}, the intersection formed is an ellipse under the condition that the angle of the plane is shallower than the slope of the pencil's edge, i.e., $|a|<|m|$. We thus have:
\begin{equation} \label{eq:majorminor}
\alpha(t) = \frac{2m\sqrt{1+a^2}d(t)}{m^2-a^2},~\beta(t) = \frac{2d(t)}{\sqrt{m^2-a^2}},
\end{equation} 
where $\{\alpha(t),\beta(t)\}$ are the major and minor axes of the elliptical cross section at $t$, respectively. We define the angle $\gamma$ such that $a = \tan(\gamma)$ which physically describes the angle in which the pencil tip is oriented against the surface. Note that when $\gamma = 0$, the pencil is pointed downward into a horizontal surface which is the perpendicular direction and $\alpha=\beta$.

\subsection{Degradation Modeling}
The degradation of the tip is modeled as function of the current state of the tip, force applied to the surface with the tip, and distance travelled while in contact. The evolution dynamics of the pencil tip can be described using $d$ from (\ref{eq:ellipse}):
\begin{equation} \label{eq:tip_model}
d(t+1) = d(t) + K_d F(t)\norm{\begin{bmatrix} x(t+1)\\y(t+1)
\end{bmatrix}-\begin{bmatrix} x(t)\\y(t)
\end{bmatrix}}_2
% (\sqrt{(x_{t+1}-x_{t})^2+(y_{t+1}-y_{t})^2}),
\end{equation} 
where $K_d$ is a scaling parameter. As the pencil is being used and wearing away, the hyperplane is pushed further into the cone. 

\subsection{Deposition Modeling}
% To determine the actual width of the deposition, $W(t)$, from a tool tip, we introduce another angle $\psi(t)$, which is the angle between the pencil's ellipsoidal cross section's major axis projected on the $x,y$ plane, and the tangent to the stroke at $t$. Using the geometry of the ellipse, we model this deposition width as:
 Using the geometry of the ellipse, we model this deposition width as:
\begin{equation}
    W(t) = \max(\alpha(t)\cos(\psi(t)), \beta(t)\sin(\psi(t))),
\end{equation}
thus the deposition width is the maximum between the major and minor axes projections measured perpendicular to the stroke at $t$.
\section{Data-Driven Control Synthesis}
In this section, we formulate our data-driven optimization approach for deposition with a degradable tool. Since the parameters from \eqref{eq:force_model} in the degradation \eqref{eq:tip_model} models are unknown, we estimate them using data from repeated strokes, \emph{before} attempting to track the reference stroke with the pencil. This is done during a training phase. 
% These repetitions are called \emph{iterations}. The value of any variable in an iteration $j$ is denoted by the superscript $(\cdot)^j$. Note that the reference stroke patterns in these offline iterations can be arbitrary, as the key purpose is parameter estimation, as explained next.

\subsection{Learning Parameters during the Training Phase}\label{sec:offline_fit}
We can rewrite \eqref{eq:force_model} as:
\begin{align}
    F(t) = \mathbf{z}^\top(t) \Theta,
\end{align}
% \begin{align}
%     F(t) = \Theta^\top \mathbf{z}(t),
% \end{align}
where $\mathbf{z}(t) = [z(t)-z_\mathrm{ref}(t), 1]^\top$ and $\Theta = [\theta, \theta_0]^\top$. In this case, the estimated parameters $\hat{\Theta}$ can be obtained using ordinary least squares as:
\begin{align}
   \hat{\Theta} = (\mathbf{Z}^\top\mathbf{Z})^{-1} \mathbf{Z}^\top \mathbf{F},
\end{align}
where
\begin{align*}
    \mathbf{Z} = \begin{bmatrix}
     z(0)-z_\mathrm{ref}(0)\\z(1)-z_\mathrm{ref}(1) \\ \vdots \\ z(T_\mathrm{off})-z_\mathrm{ref}(T_\mathrm{off})
    \end{bmatrix}, ~\mathbf{F} = \begin{bmatrix}
    F(0) \\ F(1) \\ \vdots \\ F(T_\mathrm{off})
    \end{bmatrix},
\end{align*}
where $T_\mathrm{off} \gg 0$ is the offline time used to collect data and fit the models. We then use $\hat{\Theta}$ in our optimal stroke design. Note that after learning the parameters offline, we assume the pencil is sharpened to it's initial state again, so that the given reference stroke can now be tracked using our best parameter estimates $\hat{\Theta}$.

\subsection{Optimal Stroke Tracking Synthesis}
Let $\hat{\theta}$ and $\hat{\theta}_0$ be our estimates of unknown parameters $\theta$ and $\theta_0$, respectively, obtained offline. The optimal control problem solved at $t=0$ is given by:
\begin{equation}\label{eq:optimization_problem}	\begin{array}{clll}
		\displaystyle\min_{U_t} & \displaystyle\sum\limits_{t=0}^{N} (s(t)-s_\mathrm{ref}(t))^\top Q (s(t)-s_\mathrm{ref}(t)) \\  
% 		 &~~~~~+ (y^j(t)-y^j_\mathrm{ref}(t))^\top Q_y (y^j(t)-y^j_\mathrm{ref}(t)) \\  
% 		 &~~~~~+ (D^j(t)-D^j_\mathrm{ref}(t))^\top Q_D (D^j(t)-D^j_\mathrm{ref}(t)) \Big )   \\[1ex]
% 		\text{s.t.,}  & \bar{x}_{i+1} = f_d(\bar{x}_t,u_t),\\
     	\text{s.t.,}  & \bar{x}(t+1) = A\bar{x}(t) + Bu(t),\\
         &F(t) = \hat{\theta}(z(t) - z_\mathrm{ref}(t)) + \hat{\theta}_0,\\
		&d(t+1) = d(t) + K_d F(t)\norm{\begin{bmatrix} x(t+1)\\y(t+1)
\end{bmatrix}-\begin{bmatrix} x(t)\\y(t)
\end{bmatrix}}_2\\
		&W(t) = \max(\alpha(t)\sin(\psi(t)), \beta(t)\cos(\psi(t))),\\
		&\alpha(t) = \frac{2m\sqrt{1+a^2}d(t)}{m^2-a^2}, \beta(t) = \frac{2d(t)}{\sqrt{m^2-a^2}},\\
		& \bar{x}(t) \in \mathcal{X},  u(t) \in \mathcal{U},~t = 0,1,\dots,(N-1),\\
		& s(0) = s_0,~\bar{x}(0) = \bar{x}_0,
	\end{array}
\end{equation}
where $s(t)=[x(t),y(t),W(t)]^\top$, $Q \succ 0$, and $s_0,\bar{x}_0$ are known and fixed. After computing the optimal trajectory, we apply the whole batch solution, $U_t = [u(0),u(1),...,u(N-1)]$, as an open loop rollout without re-solving \eqref{eq:optimization_problem} during the stroke. We then use an image of the drawn stroke to estimate the actual width $W_a(\zeta(t)) := W_a(t)$ along the generated stroke. \textcolor{black}{The computation of this actual width is detailed in Section~\ref{sec:exp}.} The actual and the reference stroke widths are then compared to compute an \emph{error metric} as:
\begin{equation}\label{eq:perf}
    V =  \displaystyle\sum\limits_{t=0}^{N} \vert W_a(t)-W_\mathrm{ref}(t) \vert.
\end{equation}
\textcolor{black}{Note that the reference width $W_\mathrm{ref}(t)$ is computed along the reference position trajectory for all $t$. Thus, the error metric in~\eqref{eq:perf} is only concerned with quantifying the $z$-direction force tracking error and not $x,y$ tracking. The $x,y$ position tracking error is regluated by tuning matrix $Q$ and the low level control parameters of the robot. We do not apply learning for such control loop.} 

\section{Experimental Results}\label{sec:exp}
In this section we present detailed experimental results with our proposed theory. We first show the efficacy of the offline model fitting strategy presented in Section~\ref{sec:offline_fit}, but then also highlight the iterative lowering of error metric \eqref{eq:perf} if the model parameters in \eqref{eq:force_model} are learned after iterative stroke attempts via \eqref{eq:optimization_problem}. Such iterative learning attempts also offer an intuitive bound for $T_\mathrm{off}$.

\subsection{Hardware Setup}
We perform the real-world experiments using a UR5e 6-DOF manipulator with a custom end-effector that is holding the drawing utensil. The tool is rigidly mounted, as shown in Fig.~\ref{fig:exp_space}, which imposes a hard constraint on the maximum allowable force applied by the robot arm when drawing. We use a Logitech Brio 4K Webcam to capture images of the drawings. 

\subsection{Image Processing and Computing $W_a(t)$}
In order to evaluate the closed-loop error metric of a stroke as defined in \eqref{eq:perf}, we use images of the performed stroke and compare them to a desired one. In this case, we start with a desired (i.e., reference) stroke generated from a parametric curve~\eqref{eq:parametric}. The first sequence is performed with a nominal open loop maneuver that is pre-selected. After the robot performs the stroke, the robot takes an image with the camera. The color image is converted to grayscale and then a threshold filter is applied based on the darkness of the actual stroke and the canvas. Finally, a contour filter is used to detect the stroke outline. Starting at the known beginning of the stroke and following along the parametric curve, we compute the width at each sampled point by finding the closest points which are perpendicular to the tangent line. This lets us compute the error metric~\eqref{eq:perf}.  

\subsection{Offline Force Calibration and Model Fitting}
\begin{figure}[h!]
	\centering 	
	\includegraphics[width=0.86\columnwidth]{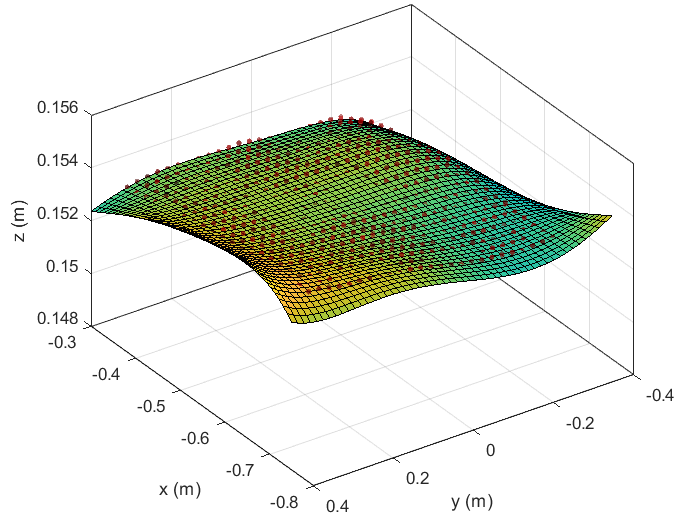}
	\caption{Surface plot of contact points on the drawing table.}
	\label{fig:force_map}    
\end{figure}

To learn the parameters in model~(\ref{eq:force_model}), we took measurements using the force sensor on the UR5e as the robot moved the tool downwards into the work surface. We ran sinusoidal sweeps of various penetration depths and used linear regression to determine the applied force as a function of penetration depth.
\begin{figure}[h!]
	\centering 	\includegraphics[width=0.9\columnwidth]{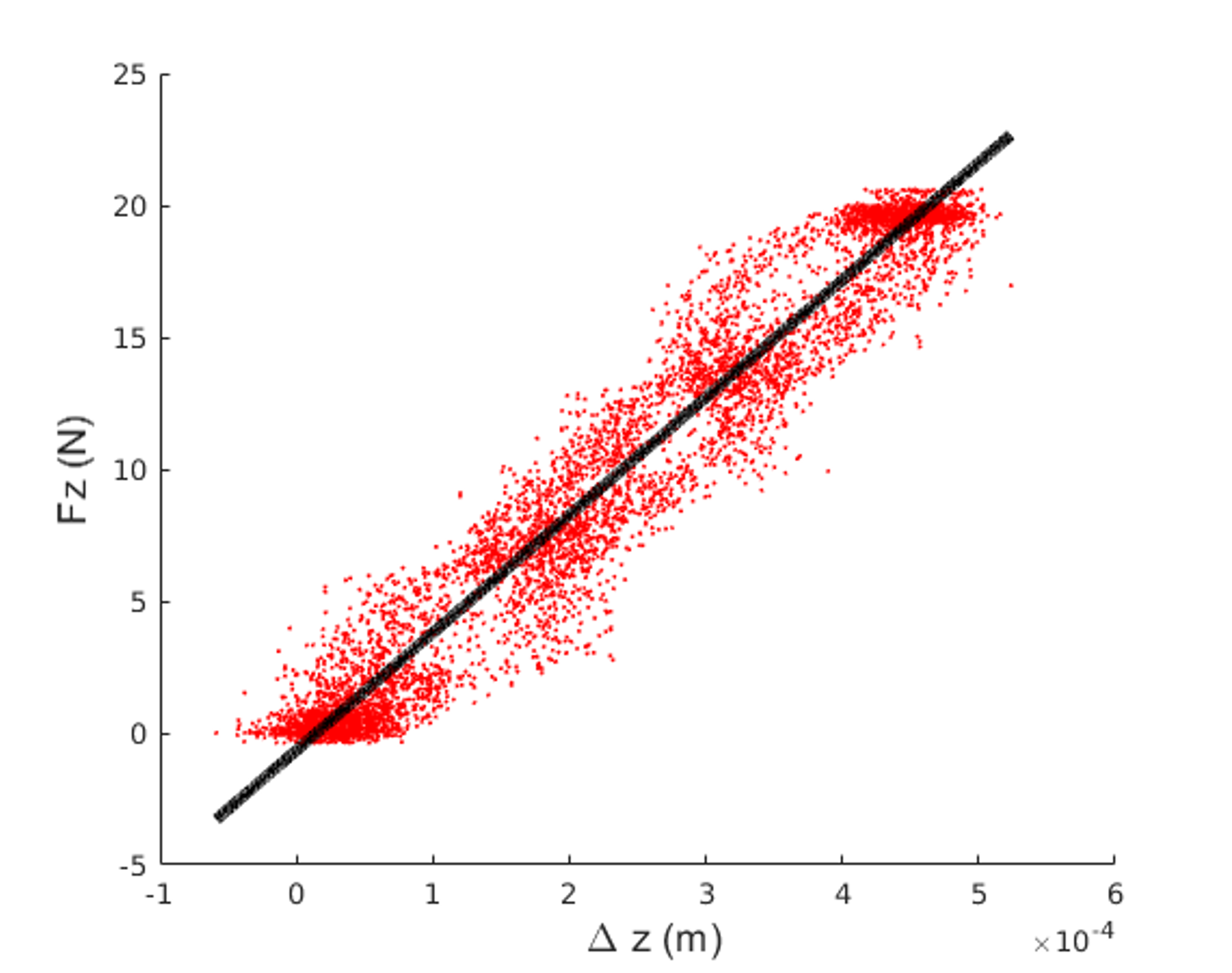}
	\caption{Force vs penetration plot.}
	\label{fig:force_calib}    
\end{figure}
Fig. \ref{fig:force_calib} shows the line of best fit plotted on top of the force measurements vs penetration depths. Due to the work surface not being flat, we recorded the contact points along the x and y directions shown in Fig. \ref{fig:force_map}. The height of the contact points are used as offsets to $z_\mathrm{ref}$ to maintain a consistent penetration depth along a stroke after contact is initially made during a trial.
\begin{figure*}[h]
\centering
    \subfloat[Iteration = 1]{%
        \includegraphics[width=0.4\columnwidth]{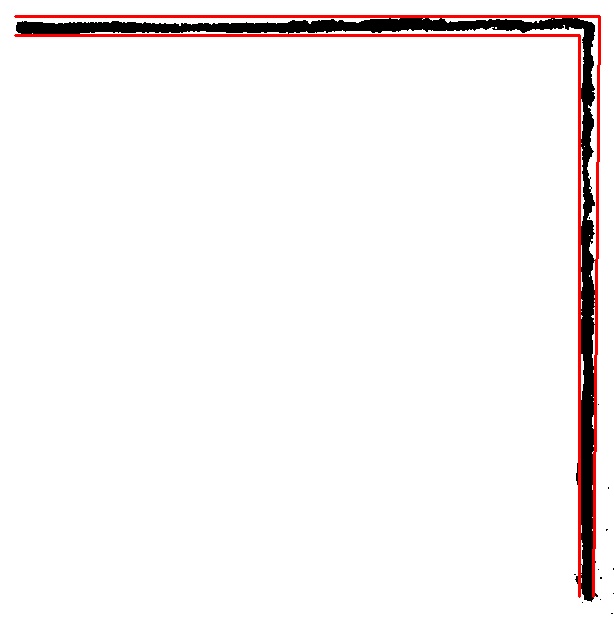}
        \label{fig:0_0}
    }
    \hspace{1em}
    \subfloat[Iteration = 2]{%
        \includegraphics[width=0.4\columnwidth]{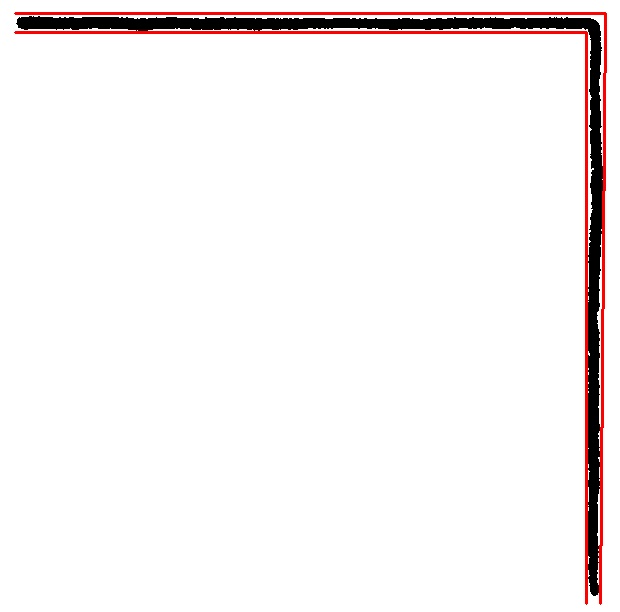}
        \label{fig:0_1}
    }
    \hspace{1em}
    \subfloat[Iteration = 5]{%
        \includegraphics[width=0.4\columnwidth]{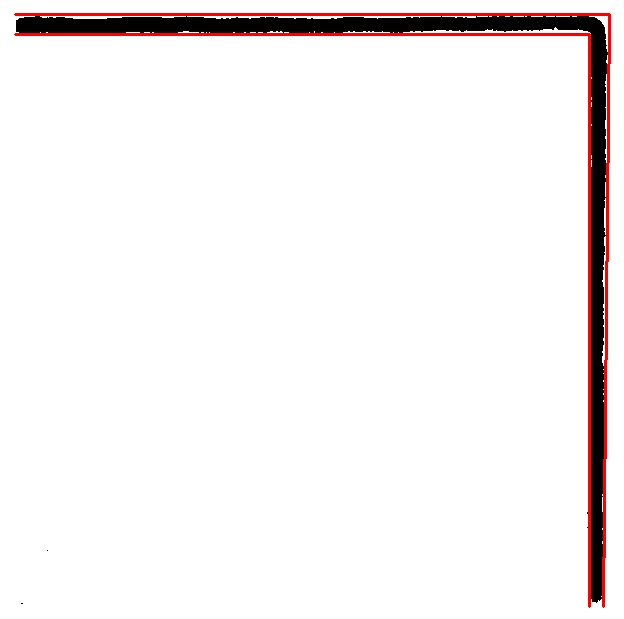}
        \label{fig:0_4}
    }
    \hspace{1em}
    \subfloat[Iteration = 10]{%
        \includegraphics[width=0.4\columnwidth]{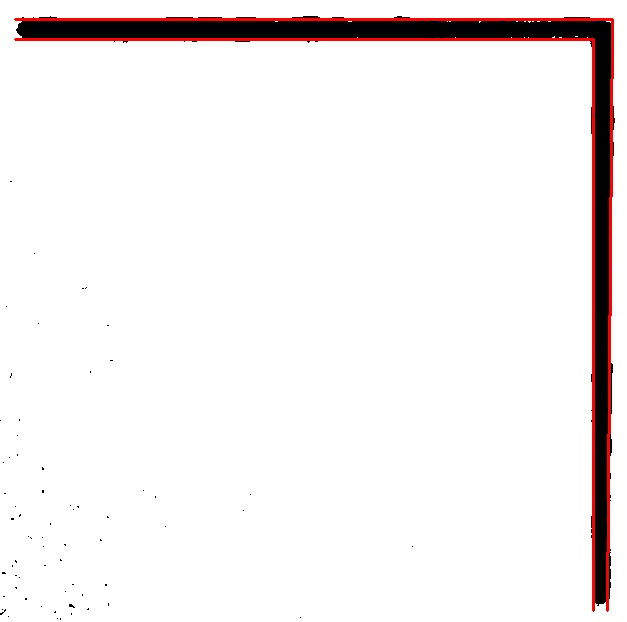}
        \label{fig:0_7}
    }
    \caption{Post-processed images of pencil strokes with $\gamma = 0^\circ$}
    \label{fig:pencil_strokes}
\end{figure*}

\begin{figure*}[h]
\centering
    \subfloat[Iteration = 1]{%
        \includegraphics[width=0.4\columnwidth]{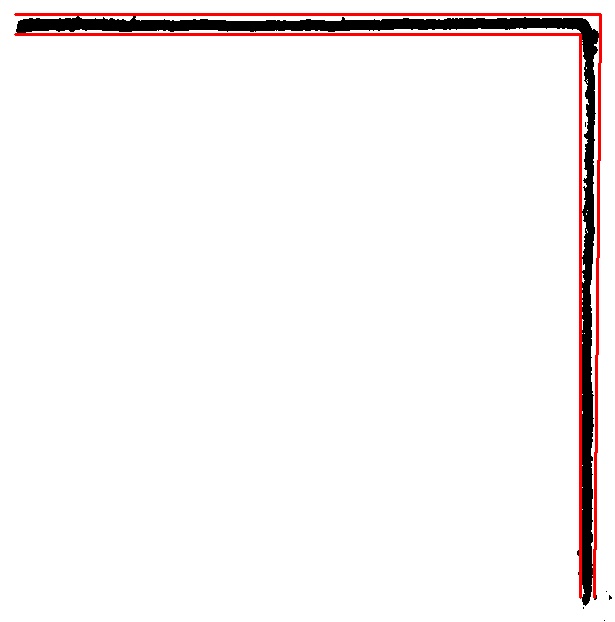}
        \label{fig:50_0}
    }
    \hspace{1em}
    \subfloat[Iteration = 2]{%
        \includegraphics[width=0.4\columnwidth]{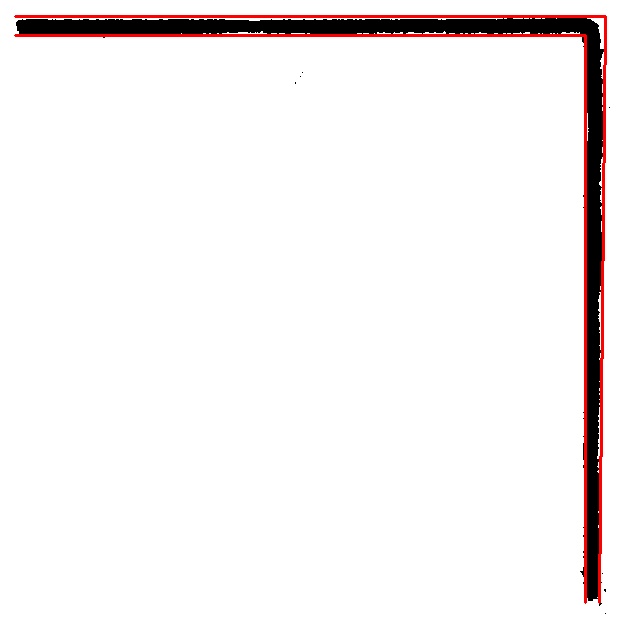}
        \label{fig:50_1}
    }
    \hspace{1em}
    \subfloat[Iteration = 5]{%
        \includegraphics[width=0.4\columnwidth]{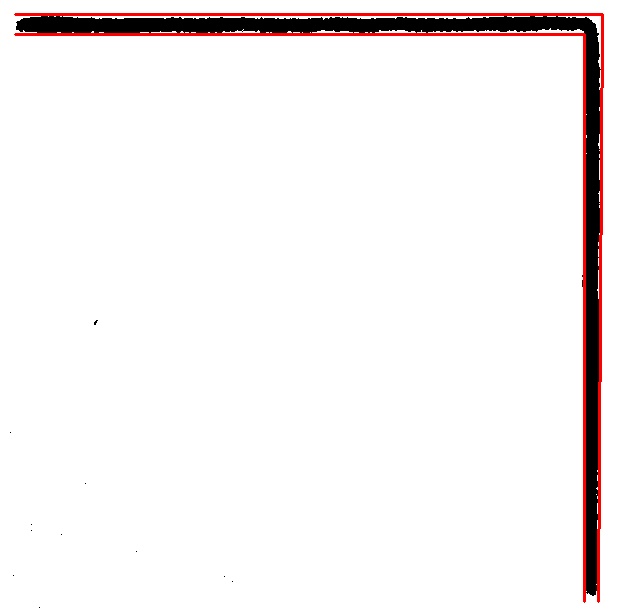}
        \label{fig:50_4}
    }
    \hspace{1em}
    \subfloat[Iteration = 10]{%
        \includegraphics[width=0.4\columnwidth]{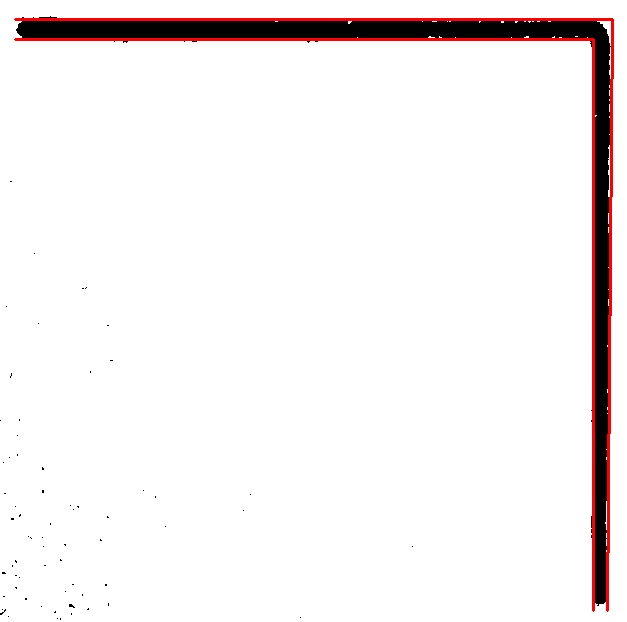}
        \label{fig:50_7}
    }\caption{Post-processed images of pencil strokes with $\gamma = 50^\circ$}
\end{figure*}

\begin{figure}[h]
	\centering 	\includegraphics[width=0.85\columnwidth]{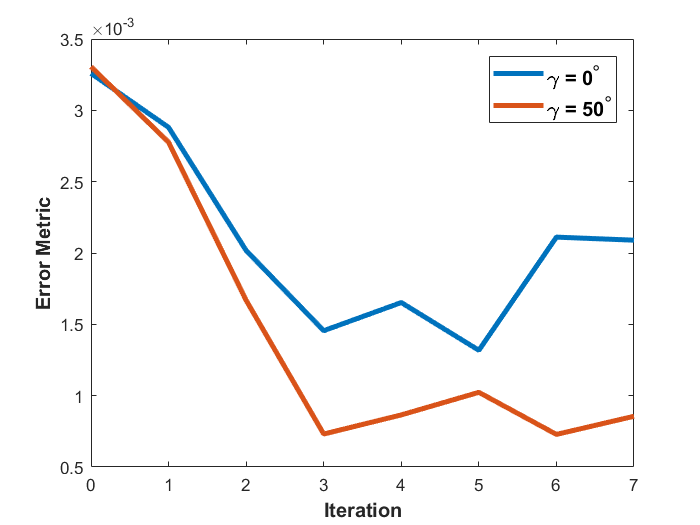}
	\caption{Comparison of the error metric with standard ($\gamma$ = 0$^\circ$) vs our approach ($\gamma$ = 50$^\circ$) .}
	\label{fig:cost_pencil}    
\end{figure}

Thus, parameters of \eqref{eq:force_model} can be learned offline as shown in Fig.~\ref{fig:force_map}. However, in the next two sections, in order to demonstrate an iterative improvement of the stroke starting from the same initial condition, instead of offline fitting as shown in Section~\ref{sec:offline_fit} and Fig.~\ref{fig:force_map}, we learn the model parameters in 
\eqref{eq:force_model} after every iterative attempt of the stroke obtained through a solution of \eqref{eq:optimization_problem}. Note, each stroke is independent, and not drawn over the previous.

\subsection{Deposition with Degradation}
We set the baseline for comparison as a pencil that is mounted perpendicular to the work surface, similar to \citep{Adamik2022,Dowd2021}, where $\gamma = 0^\circ$ and initial tip width, $\alpha(0)=\beta(0)$, as $0.1$mm. The robot performs a stroke where the desired width is $1.0$mm as it moves right and $0.7$mm as it moves down as seen in Fig.~\ref{fig:pencil_strokes}. We repeat these trials using an angled pencil where $\gamma = 50^\circ, m = 5.45, d_0 = 0.1$mm. Fig.~\ref{fig:cost_pencil} shows our approach using an angled pencil outperforms the baseline throughout the iterations with error metric lowered by about 3.6\% all the way to 65.5\%. A main reason for this is the ability to vary the deposition width by controlling the $\psi$ of the pencil while it is tilted. A pencil with $\gamma=0^\circ$ can only have a single width since the surface of the tip that makes contact with the paper is a circle. This can be a problem as seen in iteration 10 where the tip becomes larger than desired and there is no way to correct it. Our approach can also decrease the number of strokes necessary to draw a picture that requires shading since it can leverage the longer diameter while still doing precise strokes with the shorter minor axis diameter. 

\section{Conclusion}
We presented a data-driven optimization approach for robot controlled deposition with a degradable tool, specifically the robotic drawing problem. We utilized visual and force feedback to update the unknown model parameters of our tool-tip using least squares. We solved a constrained finite time optimal control problem for tracking the reference deposition profile, where our robot planned with the learned tool degradation dynamics. With real experiments on a UR5 robot, we showed that the error in target vs actual deposition decreased by up to 65\% due to the incorporation of learned degradation models in our trials. 
\balance
\bibliography{IEEEabrv,ifacconf} 
% \bibliography{ifacconf}             % bib file to produce the bibliography
                                                     % with bibtex (preferred)
                                                    
\end{document}